\documentclass[runningheads]{llncs}
\usepackage{amsmath}
\usepackage[T1]{fontenc}

\usepackage{hyperref}
\usepackage[english]{babel}
\usepackage{amsfonts}
\usepackage{amssymb}
\usepackage{url}
\usepackage{booktabs}
\usepackage{float}
\usepackage{graphicx}
\begin{document}
\title{Profiling German Text Simplification with Model-Fingerprints}
\titlerunning{Profiling German Text Simplification}
%
%
\author{Lars Klöser\orcidID{0000-0002-0931-8977} \and
Mika Elias Beele\orcidID{0000-0002-8298-2567} \and
Bodo Kraft}
\authorrunning{L. Klöser et al.}

\institute{
Aachen University of Applied Sciences, 52066 Aachen, Germany 
\email{\{kloeser,beele,kraft\}@fh-aachen.de}
}
%

\maketitle              
\begin{abstract}
While Large Language Models (LLMs) produce highly nuanced text simplifications, developers currently lack tools for a holistic, efficient, and reproducible diagnosis of their behavior. This paper introduces the Simplification Profiler, a diagnostic toolkit that generates a multidimensional, interpretable fingerprint of simplified texts. Multiple aggregated simplifications of a model result in a model's fingerprint. This novel evaluation paradigm is particularly vital for languages, where the data scarcity problem is magnified when creating flexible models for diverse target groups rather than a single, fixed simplification style. We propose that measuring a model's unique behavioral signature is more relevant in this context as an alternative to correlating metrics with human preferences. We operationalize this with a practical meta-evaluation of our fingerprints' descriptive power, which bypasses the need for large, human-rated datasets. This test measures if a simple linear classifier can reliably identify various model configurations by their created simplifications, confirming that our metrics are sensitive to a model's specific characteristics. The Profiler can distinguish high-level behavioral variations between prompting strategies and fine-grained changes from prompt engineering, including few-shot examples. Our complete feature set achieves classification F1-scores up to 71.9\%, improving upon simple baselines by over 48 percentage points. The Simplification Profiler thus offers developers a granular, actionable analysis to build more effective and truly adaptive text simplification systems.
\end{abstract}

\section{Introduction}

Automatic Text Simplification (ATS) endeavors to rephrase complex texts into more easily understandable versions to broaden information accessibility. Research in German ATS often focuses on replicating specific, fixed simplification styles inherent in reference texts \cite{rios_new_2021,toborek_new_2023,kloser_german_2024}. However, simplification must be tailored to specific circumstances to enhance the practical relevance of ATS, which includes explicitly considering the overall context. This context encompasses both the target audience\footnote{For example, children, individuals with low literacy, or domain newcomers} and other situational requirements\footnote{such as desired output length, preservation of specific contents, or an appropriate tone}.

The rise of Large Language Models (LLMs) has made generating such context-aware simplifications more feasible. For example, through straightforward prompt variations. While this flexibility is a significant advancement, it introduces a critical challenge for developers: How do we move beyond simple \textit{good/bad} scores to understand and steer the nuanced properties of these outputs? - To create solutions that are perfectly adjusted to the needs of different target groups, developers require tools that can characterize and diagnose model behavior.

Current approaches to ATS evaluation, however, do not always deliver nuanced insights in the model's behavior. The evaluation of ATS systems has long relied on reference-based metrics like BLEU \cite{papineni_bleu_2002} or SARI \cite{xu_optimizing_2016}. Reference-based metrics, by design, are inflexible. Evaluations with distinct language, domain, or simplification styles require creating novel reference simplifications to compare against. Reference-free approaches aim to avoid this need, notably learnable metrics such as REFeREE \cite{huang_referee_2024} or LENS \cite{maddela_lens_2023}.

These learnable metrics face further fundamental challenges. Firstly, there is the issue of \emph{representational breadth}. As motivated, ATS should aim for versatility. Evaluation metrics should be flexible enough to manage this versatility reliably. However, existing human evaluation datasets, if used for training or validation, typically embody only one specific, often implicit, standard of \textit{good} simplification \cite{maddela_lens_2023,alva-manchego_asset_2020,stodden_easse-_2024,naderi_subjective_2019}. Creating the multitude of high-quality, comprehensive datasets needed to cover this broad application spectrum is likely prohibitively expensive and practically unfeasible. A challenge especially for ATS in non-English languages like German, which serves as the primary linguistic focus for the empirical validation in this paper \cite{anschutz_language_2023,nagai_document-level_2024,ryan_revisiting_2023}.

Secondly, even when using the available data, there is the problem of \emph{learning reliability}. We know from other natural language processing tasks like natural language inference (NLI), question answering, or relation extraction that AI models can learn \textit{spurious patterns} – correlations present in limited data but irrelevant to the actual task \cite{chan_which_2023,glockner_breaking_2018,shinoda_which_2023,kloser_explaining_2023}. Given the limited scale of ATS data, learnable evaluation models are particularly susceptible to these pitfalls. These potentially faulty decision patterns create a high risk that these metrics will inadvertently embed biases and fail to generalize, especially when faced with diverse or out-of-domain simplifications.

Thirdly, many such learnable metrics operate as \textit{black boxes}, presenting a significant challenge in explainability. When an evaluation metric does not provide clear insights into how it arrives at a score, it becomes exceedingly difficult to diagnose its failures, understand its sensitivities, or make targeted improvements to the evaluated ATS systems. This lack of transparency harms iterative LLM development, as it provides feedback that can hardly be used for model improvements. Low scores fail to inform developers what to fix, hindering the targeted improvements necessary for creating highly adaptive ATS systems.

These issues highlight the need for a paradigm shift from monolithic evaluation to granular analysis. This paper lays the conceptual and methodological foundations for the \textit{Simplification Profiler}: a diagnostic toolkit designed to characterize ATS outputs across linguistically grounded properties. This approach provides a multi-dimensional, interpretable fingerprint of each simplification, making critical trade-offs (e.g., between simplicity and content preservation) transparent to the developer. These properties can be divided into universal quality criteria and context-dependent parameters:
\begin{itemize}
\item \emph{Universal:} Linguistic Correctness (Grammar and Spelling), Factual Accuracy (Adequacy), Coherence.
\item \emph{Context-Dependent:} Linguistic Level (Complexity), Content Scope (Compression), Terminology Use, Text Length.
\end{itemize}
The key advantage of our approach is its compositional nature. Instead of building a single, monolithic, task-specific AI model for evaluation, we leverage robust, well-established tools for specific aspects, such as NLI models, grammar checkers, and readability indices. This compositional approach allows our Profiler to stand on the shoulders of significant existing expert knowledge and broad training data, potentially increasing the individual measurements' generalizability and reliability.

These observations lead to our central hypothesis: A multi-dimensional analysis of specific text properties provides an informative and nuanced \textit{fingerprint} of ATS system performance, enabling targeted development. We test this by demonstrating that our toolkit is sensitive enough to detect the fingerprints of different development choices, using a linear classifier to distinguish outputs generated by various models and prompts. Our main technical contribution is the implementation of the toolkit and all code to reproduce the results of this paper in an open-source GitHub repository\footnote{\url{https://github.com/MSLars/German-Simplification-Profiling}}.

\section{Related Work}

\paragraph{Core Principles of ATS Evaluation.}
The evaluation of Automatic Text Simplification (ATS) targets qualities like simplicity, fluency, and adequacy \cite{shardlow_survey_2014,qiang_redefining_2025}, with recent LLM-focused work adding criteria like factuality and user engagement \cite{wu_-depth_2025,opsi_text_2024}. While human evaluation is the gold standard \cite{alfear_meta-evaluation_2024,sauberli_digital_2024}, its scalability issues necessitate automatic metrics \cite{stodden_easse-_2024}.

\paragraph{Reference-Based and -Free Metrics.}
Automatic approaches include reference-based metrics like BLEU \cite{papineni_bleu_2002}, despite its known correlation issues \cite{sulem_bleu_2018}, the ATS-specific SARI \cite{xu_optimizing_2016}, and semantic methods like BERTScore \cite{zhang_bertscore_2019}. The dependency on high-quality references spurred reference-free approaches, from traditional readability formulas to linguistically-informed structural metrics like SAMSA \cite{sulem_semantic_2018}.

\paragraph{Modern Model-Based Evaluation.}
A recent trend is the use of learnable metrics trained on human judgments (LENS \cite{maddela_lens_2023}, REFeREE \cite{huang_referee_2024}) and the direct use of \textit{LLM-as-a-Judge} \cite{gu_survey_2024,liu_evaluation_2025}, though the latter faces reliability challenges. Research into controllable generation also implies evaluating specific output properties \cite{martin_controllable_2020}. However, these approaches often still rely on datasets with a single, implicit standard of simplification quality \cite{maddela_lens_2023,alva-manchego_asset_2020,stodden_easse-_2024,naderi_subjective_2019}.

\paragraph{The Need for a German Diagnostic Toolkit.}
While comprehensive toolkits like EASSE \cite{alva-manchego_easse_2019} and its German version EASSE-DE \cite{stodden_easse-_2024} exist, they focus on aggregating overall scores. For German, property-focused analysis of novel systems often remains a manual process \cite{anschutz_language_2023,kloser_german_2024,carrer_towards_2024}. To our knowledge, a comprehensive \textit{diagnostic toolkit} for German designed to generate an \textit{interpretable fingerprint}—revealing the interplay of linguistic properties to guide development—remains an unexplored area.

\section{Methodology for Generating Test Cases}
\label{sec_simplification_method}

To validate the Simplification Profiler's diagnostic capabilities, we needed to generate a wide spectrum of simplified texts with varied characteristics. Our goal was to create a testbed reflecting common LLM development scenarios, featuring both major and subtle differences in output style. This allows us to demonstrate that our toolkit can not only distinguish between fundamentally different approaches but is also sensitive enough for fine-grained analysis, such as evaluating the impact of prompt engineering. We selected German Wikipedia as our primary data source because its encyclopedic, fact-driven style provides a controlled and standardized domain. This allows us to more clearly isolate and measure the distinct effects of different models and prompting strategies, which is the core goal of our diagnostic validation, before extending the analysis to more heterogeneous text genres.

All simplifications were generated from 5-sentence German Wikipedia excerpts using the \textit{Gemma} model family (1B, 4B, and 12B parameters), allowing us to analyze behavior at different model scales. We created our test cases using the following structured variations:

\paragraph{High-Level Behavioral Variations.}
To produce outputs with significant stylistic differences, we employed two core prompting strategies. Plain prompts provided minimal guidance and represent a common baseline approach: a generic version (\textit{plain}) and one specifying a target audience (\textit{target}). In contrast, property-oriented prompts gave explicit instructions to control specific output characteristics: one focused on maximizing content preservation (\textit{coverage}) and another on adhering to specific linguistic rules (\textit{correctness}, detailed in Appendix~\ref{sec:appendix_language_tool}). The combination of these distinct prompt strategies and model sizes creates a broad set of clearly differentiated behavioral profiles.

\paragraph{Fine-Grained Prompt Variations.}
To test the Profiler's sensitivity to more subtle changes, we created variants of the \textit{coverage} and \textit{correctness} prompts. For each of these, we generated outputs both with and without few-shot examples. This mimics a typical prompt engineering workflow and allows us to test whether our toolkit's \textit{fingerprint} is sensitive enough to detect the nuanced impact of in-context learning.

This methodology provides a rich testbed of 18 distinct simplification configurations. In our analysis, we refer to these varied approaches by the labels introduced above—distinguishing between prompt strategies (\textit{Plain}, \textit{Target}, \textit{Rules}, \textit{Content}), model sizes (\textit{1B}, \textit{4B}, \textit{12B}), and the use of few-shot samples—to investigate the unique fingerprint of each.

\section{The Simplification Profiler Toolkit}
\label{sec:eval_method}

At the heart of text simplification lies a fundamental tension of modifying the linguistic form of a text to increase simplicity while perfectly preserving its semantic content. Our Simplification Profiler is a diagnostic toolkit designed to deconstruct and quantify this tension. It provides a multidimensional fingerprint by measuring a set of well-motivated properties, each targeting an established dimension of text quality. The toolkit rests on the principle of \textit{text-ground explainability}: every score is directly derived from detectable text segments, ensuring that the resulting profile is transparent and interpretable.

The Profiler's metrics are organized into three key areas of analysis:

\subsection{Measuring Semantic Fidelity}
The most critical aspect is whether the simplification preserves the original meaning. We assess this with two NLI-based metrics.

\subsubsection{Content Correctness (COR)}
To ensure a simplification does not introduce factual errors or distort the original meaning, we measure Content Correctness. For each sentence($s_{i}^{T_O}$) in the original text ($T_O$), we use an NLI model to find the probability of a contradiction, $P(contradiction)$. We consider the complete simplification as hypothesis and test whether any $s_{i}^{T_O}$ as premise leads to a contradiction. The final score, $S_{CCor}$ , reflects the aggregated absence of contradictions.
\[ S_{CCor} = \left( \frac{1}{|T_O|} \sum_{s_{i}^{T_O} \in T_O} (1 - P(\text{contradiction})) \right) \cdot 100 \]

\subsubsection{Content Coverage (COV)}
To measure omissions, we calculate Content Coverage. The metric assesses each original meaning unit (the premise, p) against the simplification (the hypothesis, h). It combines the NLI model's entailment probability, $P(entailment)$, with a semantic similarity score, $sim(p,h)$, to reward outputs that retain the source content. This hybrid approach is robust because a high coverage score requires both strong logical entailment and high semantic similarity, making it less prone to errors than either metric would be individually.
\[ Cov(s_{i}^{T_O}) = P(\text{entailment}) \cdot sim(p,h) \]

We compute the final $S_{CCov}$ score similar to the Content Correctness.

\subsection{Measuring Linguistic Quality}
This analysis assesses the form and fluency of the simplified text.

\subsubsection{Rule-Based Simplicity (SIM) and Correctness (LNG)}
To measure adherence to simplification conventions and basic grammar, we use a set of custom rules implemented in LanguageTool (see Appendix~\ref{sec:appendix_language_tool}). Violations of simplicity rules (e.g., passive voice) decrease the SIM score, while grammatical errors (e.g., typos) decrease the LNG score. The score ($S$) is the squared harmonic mean of the non-violating word ($R_w$) and sub-clause ($R_{sc}$) ratios:
\[ S = \left(\frac{2}{\frac{1}{R_w} + \frac{1}{R_{sc}}}\right)^2 \]
This formula uses both word and sub-clause ratios to ensure the score reflects both lexical and structural rule adherence. Squaring the result more heavily penalizes any violations, making the metric more sensitive to imperfections and better distinguishing nearly-perfect outputs from those with even a few errors.

\subsubsection{Readability (FBR)}
We measure readability using a German readability index \cite{kercher_verstehen_2013}, which combines sentence and word length heuristics into a single score ($FBR_{raw}$). The detailed formula is in Appendix~\ref{sec:appendix_hix}. We normalize this score using a sigmoid function to a 0-1 scale, where higher values indicate better readability.
\[ FBR = \sigma(k \cdot (x_0 - FBR_{raw})) \]
The parameter $x_0$ centers the function, and we set it to 50 so that a raw FBR score of 50 corresponds to a normalized readability of 0.5.

\subsubsection{Coherence (COH)}
To approximate local coherence, we calculate the mean cosine similarity between the sentence embeddings of adjacent sentences. Higher scores suggest smoother semantic transitions.

\paragraph{Simple Diagnostic Heuristics.}
Alongside our core metrics, we track three simple heuristics that serve as powerful, high-level indicators of simplification style. \textbf{Text Length (LEN)} measures the output's compression relative to the source text, while the \textbf{Named Entity (ENT)} ratio serves as a proxy for the retention of key factual information. Finally, we include the \textbf{Average Sentence Length (ASL)}. While a traditional feature, our experiments (detailed in \autoref{sec:experiments}) revealed it to have disproportionately high diagnostic power for identifying specific simplification strategies. Therefore, this expressive power empirically motivates its inclusion as a key component of the fingerprint.

\subsection{Visualizing and Interpreting the Fingerprint}

The true power of the Simplification Profiler lies not in any single score but in analyzing the complete, multidimensional fingerprint. To make these profiles tangible, we visualize them as spider diagrams, which excel at revealing the unique shape and trade-offs of a simplification profile. Therefore, the visualization via spider diagrams is not merely an illustration but a core component of our methodology. It is explicitly designed to manage the complexity of the multidimensional output, allowing developers to interpret the profile holistically by its overall shape and trade-offs at a glance rather than being overwhelmed by individual numeric scores.

We applied the Profiler to a complex German source text and three distinct, generated simplifications to provide a concrete example. The English translations below illustrate the characteristics of each text:

\begin{itemize}
    \item[] \textit{"Due to the extraordinary weather conditions and the resulting agricultural production losses, renowned economic institutes are forecasting a significant increase in inflation for Germany."}
    \item[\textbf{S1:}] \textit{"Because of bad weather, experts predict higher prices."}
    \item[\textbf{S2:}] \textit{"The weather was very unusual, which is why farmers could harvest less. Well-known economic experts therefore believe that everything will become significantly more expensive."}
    \item[\textbf{S3:}] \textit{"The weather was bad, the harvests may be broen. Experts say, the moey will worth less." \footnote{The typos in this translation reflect grammatical errors present in the original German output from the Erroneous simulation.}}
\end{itemize}

\begin{figure}[t!]
    \centering
    \includegraphics[scale=0.3]{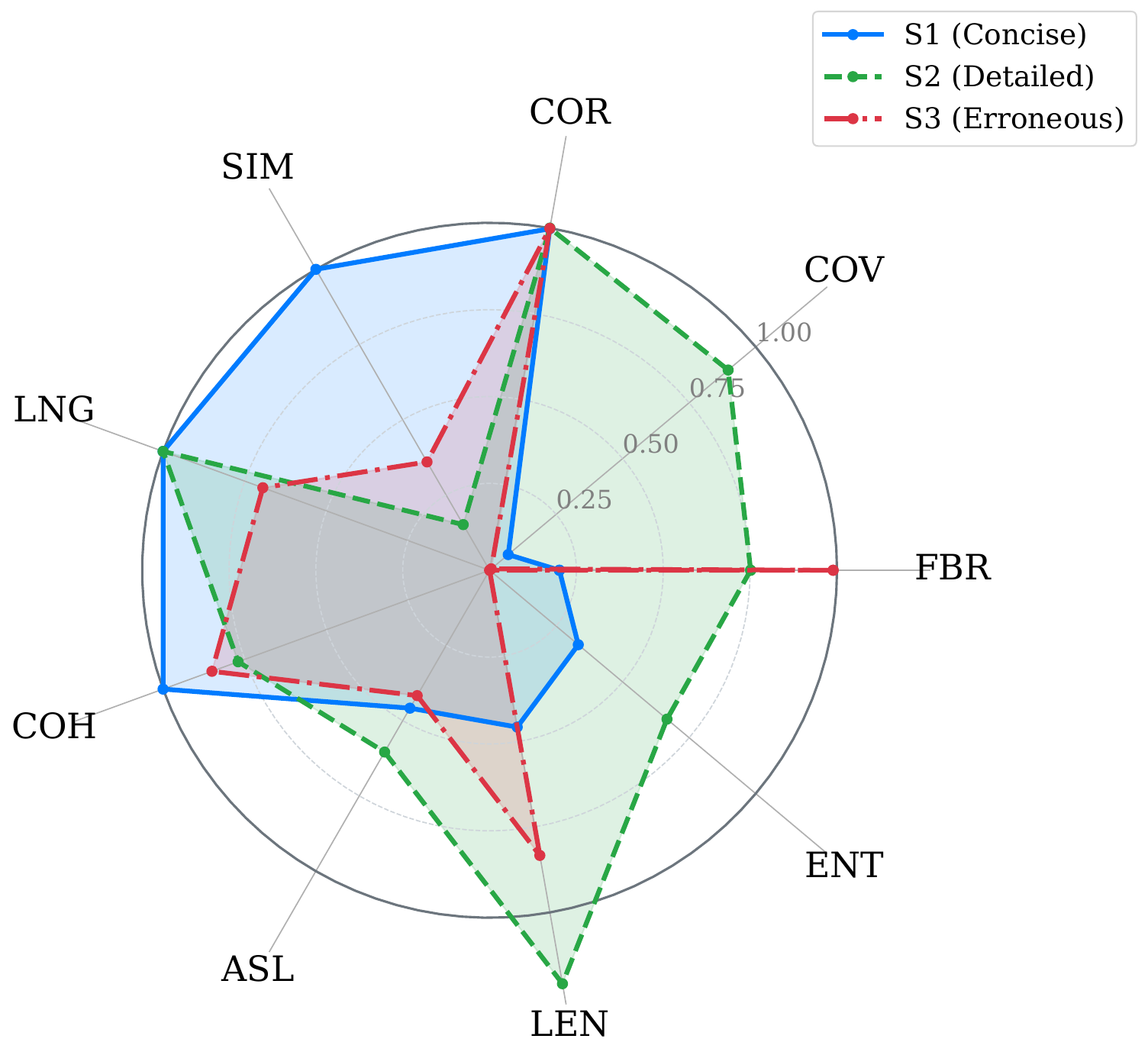}
    \caption{The fingerprints of the three example simplification strategies (S1, S2, S3). Each distinct profile shape highlights the strengths, weaknesses, and trade-offs of the corresponding simplification, such as conciseness versus content coverage.}
    \label{fig:profiler_illustration}
\end{figure}

Applying our toolkit to the original German texts yields the distinct fingerprints visualized in \autoref{fig:profiler_illustration}. The diagram makes the trade-offs immediately apparent: a \textit{Detailed} simplification (S2, dashed green) shows a high coverage, content is mostly preserved, but it uses some more sophisticated formulations. In contrast, the \textit{Concise} version (S1, solid blue) reveals a clear deficit in Content Coverage (COV), while the \textit{Erroneous} output (S3, dash-dot red) shows a dramatic failure in Linguistic Correctness (LNG).

\section{Evaluation Data}
\label{sec:eval_data}
Our analysis is reference-less, meaning our Profiler does not require pre-existing gold standard simplifications for evaluation. This section details the data-driven methodology we used to construct a compact yet representative evaluation subset from our primary data source, the German Wikipedia dataset\footnote{The German portion of the Wikipedia dataset is available at \url{https://huggingface.co/datasets/wikimedia/wikipedia}}. The goal was to build a challenging testbed specifically designed to probe the fine-grained diagnostic capabilities of our toolkit.

\subsection{Construction of the Evaluation Subset}
\label{ssec:eval_subset_construction}

A robust diagnostic toolkit requires a challenging and diverse test dataset. Our initial analysis of German Wikipedia revealed a highly skewed distribution of simplification phenomena, as detailed in \autoref{tab:rule_distribution}. This analysis confirmed that random sampling would fail to capture many rare but important linguistic challenges. We employed a greedy sampling algorithm to build a benchmark that explicitly contains these phenomena. This data-driven approach allowed us to construct a compact (~1,000 examples) yet diverse dataset with a guaranteed high concentration of various simplification targets, making it ideal for fine-grained analysis. Each example consists of a five-sentence window, with sentences segmented using spaCy and filtered to ensure grammatical completeness (e.g., containing a verb and proper terminal punctuation).

\begin{table}[ht]
\centering
\caption{Distribution of selected comprehensibility rule violations across 10,000 excerpts randomly sampled from German Wikipedia. Each excerpt consists of five consecutive sentences from a randomly selected article. The imbalance in rule frequency illustrates the need for targeted sampling to ensure coverage of rare but relevant linguistic phenomena.}
\label{tab:rule_distribution}
\begin{tabular}{@{}l r@{}}
\toprule
\textbf{Rule} & \textbf{Share (\%)} \\
\midrule
\multicolumn{2}{l}{\textbf{Top 3 most frequent rule violations}} \\
\addlinespace[4pt]
Sentence length       & 84.71\% \\
Long words            & 72.99\% \\
Abstract words        & 70.68\% \\
\midrule
\multicolumn{2}{l}{\textbf{Top 3 least frequent rule violations}} \\
\addlinespace[4pt]
Anglicisms            & 12.57\% \\
Negation              & 10.16\% \\
Technical terms       & 3.37\% \\
\bottomrule
\end{tabular}
\end{table}

\section{Experimental Validation}
\label{sec:experiments}

To validate that the Simplification Profiler can generate meaningful and diagnostically useful fingerprints, we designed a series of classification experiments. The central hypothesis is that if the metric profiles are sufficiently sensitive, a simple linear classifier should be able to distinguish texts produced by different models and prompting strategies reliably. This section details this validation, moving from high-level distinctions to fine-grained analysis.

\subsection{Experimental Setup}

\paragraph{Data and Configurations.} Our dataset consists of 14,868 simplified texts generated from 826 unique source excerpts using the 18 distinct configurations described in \autoref{sec_simplification_method}.

\paragraph{The Classification Task.} For our main validation, we frame the problem as a series of binary classification tasks (e.g., \textit{Was this simplification created with target prompt?} or \textit{Was this text generated by the 1B model?}). Success in these tasks provides direct evidence of the Profiler's descriptive power.

\paragraph{Features and Baselines.} The feature vector for each simplification comprises the 23 metrics from the Profiler toolkit. We compare the performance of a \textit{LogisticRegression} classifier trained on this full feature set against two baselines: a `Random` classifier and a \textit{Simple Baseline} trained only on six basic length-related features. This allows us to quantify the added value of our comprehensive linguistic fingerprint.

\subsection{Results: Diagnostic Sensitivity of the Fingerprints}

A preliminary analysis reveals apparent differences in the average metric scores between strategies. For instance, \textit{Target} prompts produce longer texts with higher FBR scores than \textit{Plain} prompts, suggesting the existence of distinct profiles. Our classification experiments formally test this.

\paragraph{High-Level Distinguishability.}
\autoref{tab:model_eval_summary_cv} shows the main classification results. Our Profiler's complete feature set enables the classifier to outperform both baselines across nearly all categories significantly. The dramatic F1-score improvement for \textit{Target} prompts (48.3\% vs. 0.0\% for the Simple Baseline) is particularly notable, confirming that the rich linguistic fingerprint captures nuances that simple length metrics miss entirely. The strong overall performance validates that the Profiler can reliably distinguish significant behavioral differences between models and prompt archetypes.

\begin{table}
\centering
\caption{The diagnostic sensitivity of the Simplification Profiler. A linear classifier using the Profiler's rich linguistic metrics significantly outperforms simple baselines in identifying text origins , confirming the metrics provide a distinct and valuable fingerprint.}
\label{tab:model_eval_summary_cv}
\resizebox{\textwidth}{!}{
\begin{tabular}{@{}lcccc@{}}
\toprule
Target Label & Accuracy ($\pm$ std) & F1-Score ($\pm$ std) & Simple Baseline 
(Acc/F1 $\pm$ std) & Random Baseline (Acc/F1 $\pm$ std)\\
\midrule
Plain & 84.4 $\pm$ 0.7 & 39.1 $\pm$ 1.9 & (83.7 $\pm$ 0.7 / 35.2 $\pm$ 2.1) 
& (71.9 $\pm$ 0.9 / 16.9 $\pm$ 2.0) \\
Target & 86.6 $\pm$ 0.4 & 48.2 $\pm$ 1.7 & (83.2 $\pm$ 0.0 / 0.0 $\pm$ 0.0) 
& (71.8 $\pm$ 0.2 / 16.4 $\pm$ 1.4) \\
Rules & 78.9 $\pm$ 0.5 & 64.6 $\pm$ 1.1 & (78.0 $\pm$ 0.8 / 62.4 $\pm$ 1.6) 
& (55.6 $\pm$ 0.5 / 33.6 $\pm$ 0.5) \\
Content & 80.7 $\pm$ 0.5 & 67.1 $\pm$ 1.0 & (78.0 $\pm$ 0.6 / 61.1 $\pm$ 
1.2) & (55.6 $\pm$ 0.6 / 33.5 $\pm$ 0.9) \\
1b & 81.7 $\pm$ 0.2 & 71.9 $\pm$ 0.3 & (77.7 $\pm$ 0.5 / 65.7 $\pm$ 0.7) & 
(54.8 $\pm$ 0.6 / 32.4 $\pm$ 0.9) \\
4b & 66.4 $\pm$ 0.2 & 3.4 $\pm$ 1.1 & (66.6 $\pm$ 0.0 / 0.0 $\pm$ 0.0) & 
(55.2 $\pm$ 0.5 / 32.9 $\pm$ 0.8) \\
12b & 80.1 $\pm$ 1.1 & 67.0 $\pm$ 1.8 & (78.6 $\pm$ 1.7 / 63.5 $\pm$ 3.0) & 
(55.9 $\pm$ 0.6 / 33.9 $\pm$ 1.0) \\
\bottomrule
\end{tabular}}
\end{table}

\paragraph{Fine-Grained Analysis.}
We conducted two further experiments to probe the Profiler's sensitivity. First, to investigate the poor performance for the 4b class in \autoref{tab:model_eval_summary_cv}, we performed pairwise classification between model sizes (\autoref{tab:classification_1B_4B_12B}). The high success rate in these head-to-head tests confirms the fingerprints are distinct and that the initial poor result was an artifact of the linear classifier's inability to handle the \textit{middle-ground} problem, further highlighting the Profiler's diagnostic utility.

Second, as an ultimate sensitivity test, we checked if the Profiler could detect the subtle impact of few-shot prompting. As shown in \autoref{tab:classification_FS_no_FS}, the classifier successfully distinguishes between the \textit{FS} and \textit{NoFS} variants for both \textit{Rules} and \textit{Content} prompts. The successful classification confirms the Profiler is sensitive enough for fine-grained prompt engineering analysis.

\begin{table}
\label{tab:classification_1B_4B_12B}
\centering
\caption{Pairwise classification results between model sizes. The high head-to-head accuracy confirms each model size has a distinct fingerprint, indicating the \textit{middle-ground} problem for the 4B model in the general classification task was a classifier artifact.}
\resizebox{\textwidth}{!}{
\begin{tabular}{@{}lcccc@{}}
\toprule
Target Labels & Accuracy ($\pm$ std) & F1-Score ($\pm$ std) & Simple Baseline 
(Acc/F1 $\pm$ std) & Random Baseline (Acc/F1 $\pm$ std)\\
\midrule
1B vs. 4B & 80.25 $\pm$ 0.3 & 79.77 $\pm$ 0.5 & (77.60 $\pm$ 1.0 / 76.68 $\pm$ 1.3) & (4997 $\pm$ 0.4 / 4974 $\pm$ 0.5) \\
4B vs. 12B & 68.96 $\pm$ 0.7 & 68.96 $\pm$ 0.7 & (65.94 $\pm$ 1.1 / 66.00 $\pm$ 1.1) & (49.97 $\pm$ 0.4 / 49.74 $\pm$ 0.5) \\
1B vs. 12B & 86.57 $\pm$ 0.0 & 86.04 $\pm$ 0.4 & (85.59 $\pm$ 0.4 / 84.65 $\pm$ 0.6) & (49.97$\pm$ 0.4 / 49.74 $\pm$ 0.5) \\

\bottomrule
\end{tabular}}
\end{table}

\begin{table}
\label{tab:classification_FS_no_FS}
\centering
\caption{The Profiler's sensitivity to fine-grained changes is confirmed, as its metric fingerprint allows a classifier to distinguish outputs generated with and without few-shot prompting.}
\resizebox{\textwidth}{!}{
\begin{tabular}{@{}lcccc@{}}
\toprule
Target Labels & Accuracy ($\pm$ std) & F1-Score ($\pm$ std) & Simple Baseline 
(Acc/F1 $\pm$ std) & Random Baseline (Acc/F1 $\pm$ std)\\
\midrule
Rules & 63.22 $\pm$ 1.5 & 61.77 $\pm$ 1.5 & (59.54 $\pm$ 1.7 / 58.67 $\pm$ 2.1) & (50.32 $\pm$ 1.9 / 50.34 $\pm$ 2.0) \\
Content & 77.30 $\pm$ 1.6 & 77.30 $\pm$ 1.5 & (76.49 $\pm$ 1.1 / 76.39 $\pm$ 1.1) & (50.32 $\pm$ 1.9 / 50.34 $\pm$ 2.0) \\
\bottomrule
\end{tabular}}
\end{table}

\subsection{Visualizing and Interpreting the Fingerprints}
\begin{figure}
    \centering
    \includegraphics[scale=.3]{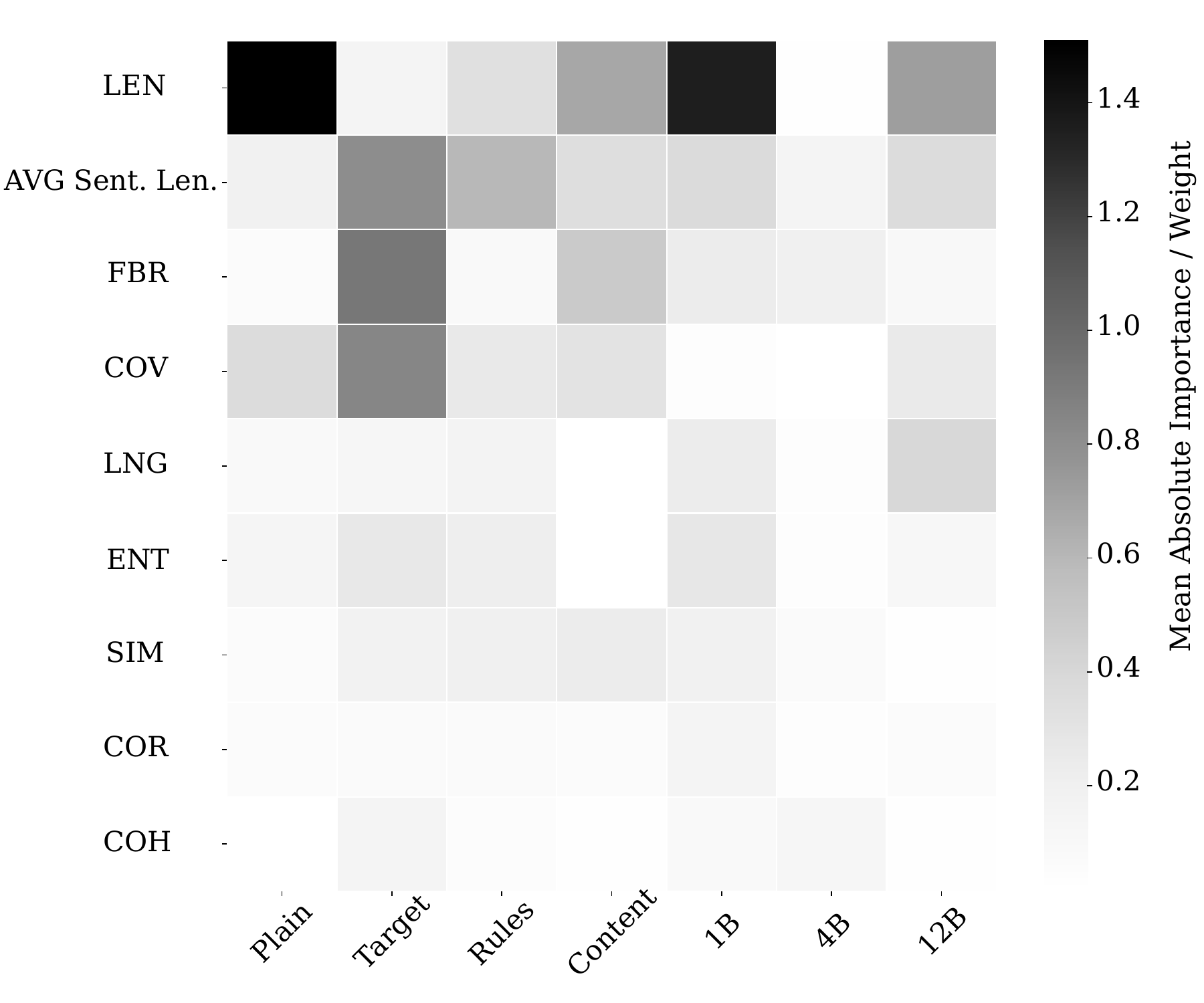}
    \caption{Feature importance heatmap showing the distinct metric patterns for identifying each condition. For example, the \textit{Target} prompt is strongly characterized by readability (FBR) and content coverage (COV), while the \textit{1B} model is primarily identified by its length (LEN).}
    \label{fig:feature_weights}
\end{figure}

We can visualize the feature importance to understand why these fingerprints are so effective (\autoref{fig:feature_weights}). The heatmap clearly shows that different strategies have different key identifiers. For instance, \textit{Target} prompts are defined by FBR and COV, while LEN defines 1B model outputs. The ablation study in \autoref{tab:avg_sent_length_importance} quantifies this, showing that removing Average Sentence Length impedes the ability to detect Target and Rules prompts, confirming the critical and sometimes non-obvious role of simple heuristics within a complex fingerprint.

\begin{table}
\centering
\caption{Ablation study for Average Sentence Length. Removing this single feature significantly harms classifier performance for \textit{Target} and \textit{Rules} prompts, confirming its critical role in characterizing these simplification strategies.}
\label{tab:avg_sent_length_importance}
\resizebox{\textwidth}{!}{
\begin{tabular}{@{}lcc@{}}
\toprule
Target Label & Full Feature Set (Acc/F1 $\pm$ std) & Without Avg. Sent. Len. (Acc/F1 $\pm$ std)\\
\midrule
Plain & 84.4 $\pm$ 0.6 / 39.1 $\pm$ 1.7 & 84.2 $\pm$ 0.6 / 38.2 $\pm$ 1.9 \\
Target & 86.6 $\pm$ 0.4 / \textbf{48.3 $\pm$ 1.7} & 84.1 $\pm$ 0.5 / \textbf{24.9 $\pm$ 2.1}\\
Rules & 78.9 $\pm$ 0.5 / \textbf{64.6 $\pm$ 1.1} & 73.9 $\pm$ 0.2 / \textbf{52.1 $\pm$ 0.4} \\
Content & 80.7 $\pm$ 0.5 / 67.1 $\pm$ 1.0 & 80.8 $\pm$ 0.6 / 67.4 $\pm$ 1.1 \\
1B & 81.7 $\pm$ 0.2 / 71.9 $\pm$ 0.3 & 81.5 $\pm$ 0.4 / 71.5 $\pm$ 0.6 \\
4B & 66.4 $\pm$ 0.2 / 3.4 $\pm$ 1.0 & 66.6 $\pm$ 0.1 / 0.6 $\pm$ 0.4 \\
12B & 80.1 $\pm$ 1.1 / 67.0 $\pm$ 1.9 & (8.8 $\pm$ 0.8 / 64.7 $\pm$ 1.3 \\
\bottomrule
\end{tabular}}
\end{table}

\begin{figure}[H]
    \centering
    \includegraphics[width=0.8\textwidth]{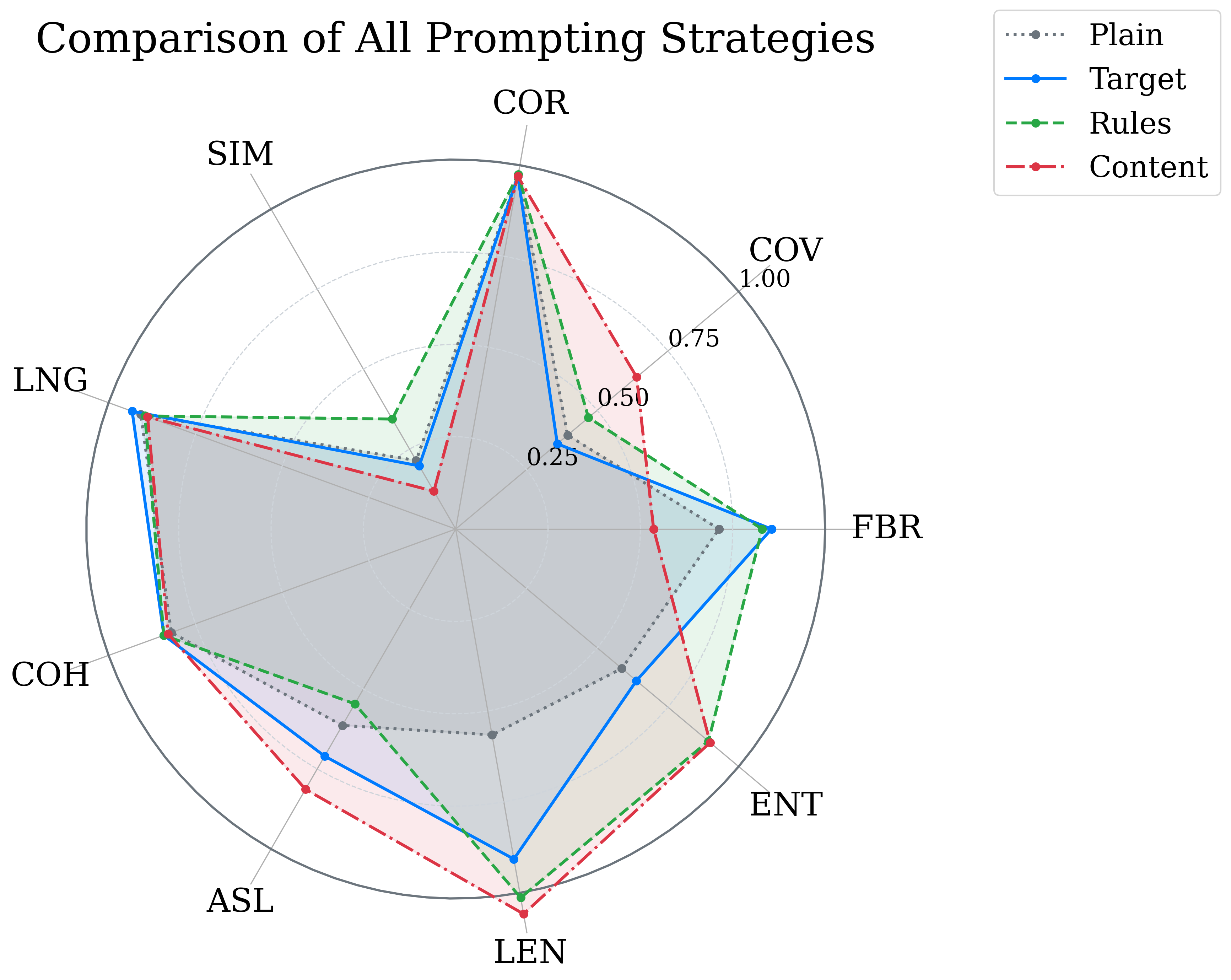}
    \caption{Overlaid fingerprints of the four main prompting strategies. The distinct shape of each profile highlights the inherent trade-offs of each strategy, such as the focus on readability versus content coverage.}
    \label{fig:fingerprint_overlay}
\end{figure}

While the feature importance heatmap in \autoref{fig:feature_weights} reveals which individual metrics are most discriminative, the overlay spider diagram in \autoref{fig:fingerprint_overlay} provides a holistic, qualitative view of the archetypal fingerprint for each prompting strategy. The visual differences are striking. For example, the \textit{Target} profile (blue, solid line) skews towards high Readability (FBR) but at the cost of lower Content Coverage (COV), while the \textit{Content} profile (red, dash-dot line) shows the opposite trade-off. Furthermore, the plot reveals non-obvious relationships between metrics. One might assume that a lower Average Sentence Length (ASL) would directly lead to the highest FBR score, but the chart shows otherwise. The \textit{Rules} prompt (green, dashed line) produces the shortest sentences, yet the \textit{Target} prompt achieves superior readability, suggesting it simplifies via other means like vocabulary choice rather than sentence splitting. This direct visual comparison of strategic trade-offs is the core diagnostic power of the Simplification Profiler.

\section{Limitations}
\paragraph{Scope of Analysis: A Diagnostic Tool, Not a Human Proxy.}
The Simplification Profiler primarily serves as a \textit{diagnostic toolkit for differential analysis}—that is, it helps developers understand the effects of their changes by comparing the fingerprints of System A versus System B. It does not substitute for holistic human quality judgment but is a complementary tool for the iterative development phase. Our goal is to empower developers to diagnose \textit{how} their design choices (e.g., a new prompt) impact the output's characteristics \textit{before} undertaking costly and time-consuming human evaluation.

Furthermore, a single human preference score becomes less meaningful as ATS systems become more adaptive. The goal is to produce varied outputs for different audiences (e.g., one optimized for conciseness, another for content preservation). Our toolkit allows developers to deliberately craft and verify these different profiles. Establishing the correlation of these multi-faceted fingerprints with human judgments for specific user groups is a crucial but distinct subsequent step in the development pipeline and a valuable area for future work.

Our toolkit's compositional design means its effectiveness is inherently tied to the performance of its constituent components. For instance, the Content Correctness (COR) and Content Coverage (COV) metrics rely entirely on an NLI model. Any biases, domain gaps, or inaccuracies in this NLI model will propagate directly into our evaluation scores. Similarly, the COH (Coherence) metric, based on the cosine similarity of sentence embeddings, is a heuristic for local coherence and may not capture more complex, discourse-level coherence phenomena or penalize valid simplifications that intentionally restructure sentence flow. The SIM and LNG scores are also limited by the scope and accuracy of the rules implemented in the underlying LanguageTool checker.

Our validation was limited to German Wikipedia articles. Porting the framework is a non-trivial effort due to its language-specific tools (e.g., FBR, German LanguageTool rules), and applying it to new domains may require different evaluation dimensions to account for their unique simplification characteristics. Therefore, the immediate generalizability of our findings and tool to other languages and domains is limited.

While our framework covers core aspects of simplification quality, such as linguistic correctness, adequacy, and complexity, the evaluated dimensions are not exhaustive. For specific applications, other attributes may be equally or more important. For example, our current metrics do not explicitly assess the appropriateness of the text's tone or register, nor do they capture aspects of user engagement or trust. A comprehensive evaluation might need to incorporate metrics for these more pragmatic qualities, depending on the specific simplification goal.

\section{Conclusions}

The increasing sophistication of Large Language Models in Automatic Text Simplification has exposed a critical gap: developers lack the tools for a holistic, efficient, and reproducible diagnosis of model behavior. This paper argued for a paradigm shift toward granular diagnostic analysis as an alternative to costly and often inflexible human preference studies. We introduced the Simplification Profiler, a diagnostic toolkit designed to provide a multi-dimensional, interpretable fingerprint of any simplified text. This approach is particularly vital to overcoming the problem of data scarcity in German ATS and for developing flexible models for diverse target groups where data scarcity makes traditional evaluation methods impractical.

Our empirical evaluation systematically tested our toolkit by generating simplifications across various models and prompting strategies. The results demonstrated that these different development choices leave distinct and measurable signatures on the output texts. We showed that the Simplification Profiler is sensitive enough to capture these differences. A simple linear classifier could reliably distinguish the originating experimental condition based solely on the generated fingerprint. These findings confirm our central hypothesis: a multi-dimensional analysis of specific text properties provides an informative and nuanced fingerprint of ATS system performance, enabling targeted development.

Our work's primary implication is enabling a more principled and efficient development and analysis for ATS. The Simplification Profiler allows moving beyond simple trial and error by making the trade-offs between properties like readability, content coverage, and linguistic correctness transparent. Researchers and practitioners diagnose the specific effects of optimizations, understand model behaviors, and make targeted adjustments to create ATS solutions that are truly adaptive to the varied needs of different target groups. Our main technical contribution, the open-source implementation of our toolkit, serves as a practical first step in this direction.

While this work establishes the Profiler's diagnostic utility, it also lays the groundwork for future research. The path forward involves addressing the current limitations by conducting human correlation studies, expanding to new languages, and enriching the fingerprint with more nuanced properties like tone and register. The evolution of such diagnostic toolkits is a key step towards creating an ecosystem where developers can craft text simplification systems with greater precision, responsibility, and awareness of their target users.

\bibliographystyle{splncs04}
\bibliography{EXPLAINS}

\appendix

\section{LanguageTool Rule Overview}
\label{sec:appendix_language_tool}

\begin{table}[H]
\centering
\caption{Overview of LanguageTool community‐edited rule categories. Grammar rules cover general spelling, punctuation, and usage issues, while style rules are often highly specific to particular phrases or terms.}
\label{tab:languagetool_rules}
\begin{tabular}{@{}l p{0.65\linewidth}@{}}
\toprule
\textbf{Name} & \textbf{Description} \\
\midrule
\multicolumn{2}{l}{\textbf{Examples for grammar rules}} \\ 
\addlinespace[4pt]
Typos                          & Detects potential misspellings and simple typing errors. \\
Confused words                 & Flags words that look or sound similar and are often mixed up (e.g.\ \emph{plant} vs.\ \emph{planet}). \\
Capitalization                 & Checks for incorrect uppercase/lowercase usage (e.g.\ nominalizations, sentence starts). \\
Correspondence                 & Validates greetings and sign‐off formats in letters and emails. \\
Idioms                         & Ensures correct use of fixed expressions (e.g.\ \emph{in droves} vs.\ \emph{in streams}). \\
\midrule
\multicolumn{2}{l}{\textbf{Examples for style rules}} \\ 
\addlinespace[4pt]
Style suggestions              & Broad category covering clarity, tone, redundancy, and formality. \\
Adjective→adverb               & Recommends replacing rare adjective forms (e.g.\ \emph{previous}) with adverbial forms (e.g.\ \emph{previously}). \\
Preposition in phrase          & Corrects fixed‐phrase preposition use (e.g.\ \emph{to note} → \emph{for noting}). \\
Name consistency               & Enforces standardized spelling of specific names and brands (e.g.\ \emph{OpenStreetMap}). \\
Term standardization  & Validates and corrects specialized terms (e.g.\ \emph{JavaScript applet}). \\
\bottomrule
\end{tabular}
\end{table}

\begin{table}[H]
\centering
\caption{Overview of LanguageTool community‐edited \textit{understandability} rule categories. Sentence‐level rules target clarity by avoiding complex structures, while word‐level rules discourage difficult or technical vocabulary.}
\label{tab:languagetool_understandability}
\begin{tabular}{@{}l p{0.65\linewidth}@{}}
\toprule
\textbf{Name} & \textbf{Description} \\
\midrule
\multicolumn{2}{l}{\textbf{Sentence‐level rules}} \\ 
\addlinespace[4pt]
Sentence length            & Warns when a sentence exceeds recommended length thresholds. \\
Perfect tense              & Prefers present perfect over simple past for easier comprehension. \\
Negation                   & Advises removing negations to simplify statements. \\
Subordinate clauses        & Suggests breaking up or removing subordinate clauses. \\
Relative clauses           & Encourages avoiding relative clauses for direct phrasing. \\
Passive voice              & Flags passive constructions; promotes active voice. \\
Genitive case              & Recommends replacing genitive phrases with simpler constructions (e.g.\ “of the”). \\
One idea per sentence      & Ensures each sentence contains only one main idea. \\
Subjunctive mood           & Discourages subjunctive/indirect speech for clarity. \\
\midrule
\multicolumn{2}{l}{\textbf{Word‐level rules}} \\ 
\addlinespace[4pt]
Abstract words             & Recommends concrete terms over abstract nouns. \\
Long words                 & Warns when individual words exceed recommended length. \\
\bottomrule
\end{tabular}
\end{table}

\section{Details of the FBR index}
\label{sec:appendix_hix}

The FBR heuristic is calculated from six component indices, which are first normalized and then aggregated. The components are divided into two categories:

\subsubsection*{Sentence Complexity Indices}
\begin{enumerate}
    \item \textbf{S1: Average Sentence Length in Syllables}
    \[ S_1 = (x_{S1} - 12.37) \cdot \frac{100}{24.12 - 12.37} \]
    where $x_{S1}$ is the average sentence length in syllables.

    \item \textbf{S2: Sentences with > 6 Words}
    \[ S_2 = (x_{S2} - 41.77) \cdot \frac{100}{67.42 - 41.77} \]
    where $x_{S2}$ is the percentage of sentences with more than 6 words.

    \item \textbf{S3: Sentences with > 16 Words}
    \[ S_3 = (x_{S3} - 22.10) \cdot \frac{100}{52.59 - 22.10} \]
    where $x_{S3}$ is the percentage of sentences with more than 16 words.

    \item \textbf{S4: Sentences with > 20 Words}
    \[ S_4 = (x_{S4} - 21.90) \cdot \frac{100}{64.67 - 21.90} \]
    where $x_{S4}$ is the percentage of sentences with more than 20 words.
\end{enumerate}

\subsubsection*{Word Complexity Indices}
\begin{enumerate}
    \setcounter{enumi}{4} 
    \item \textbf{W1: Average Word Length in Syllables}
    \[ W_1 = (x_{W1} - 1.936) \cdot \frac{100}{2.339 - 1.936} \]
    where $x_{W1}$ is the average word length in syllables.

    \item \textbf{W2: Words with > 3 Syllables}
    \[ W_2 = (x_{W2} - 10.75) \cdot \frac{100}{21.21 - 10.75} \]
    where $x_{W2}$ is the percentage of words with more than 3 syllables.
\end{enumerate}

\subsubsection*{Final Score Calculation}
The component indices are first averaged into a Sentence Complexity Score ($K_S$) and a Word Complexity Score ($K_W$):
\[ K_S = \frac{S_1 + S_2 + S_3 + S_4}{4} \quad \text{and} \quad K_W = \frac{W_1 + W_2}{2} \]
The final index is the mean of these two scores: $FBR_{raw} = (K_S + K_W) / 2$.
\end{document}